\documentclass[runningheads]{llncs}
\usepackage{graphicx}
\usepackage{amssymb}
\newcommand{\field}[1]{\mathbb{#1}}
\newcommand{\R}{\field{R}}
\newcommand{\N}{\field{N}} 
\newcommand{\E}{\field{E}}
\usepackage{mathtools}

\usepackage{subfig}

\usepackage{pgfplots}  
\pgfplotsset{compat=newest}

\usepackage{filecontents}
\usepackage{pgfplots, pgfplotstable}
\usepgfplotslibrary{statistics}

\usepackage{rotating,booktabs,multirow}

\usepackage{amssymb}

\begin{document}
\title{Reputation-driven Decision-making in
Networks of Stochastic Agents}
%
%
\author{
David Maoujoud\inst{1}
\and
Gavin Rens\inst{2}
}
\authorrunning{D. Maoujoud and G. Rens}
%
\institute{
Katholieke Universiteit Leuven, Belgium\\
\email{david.maoujoud@student.kuleuven.be},
\email{david.maoujoud@hotmail.com}
\and
Katholieke Universiteit Leuven, Belgium\\
\email{gavin.rens@kuleuven.be}
}
\maketitle              
\begin{abstract}

This paper studies multi-agent systems that involve networks of self-interested agents. We propose a Markov Decision Process-derived framework, called RepNet-MDP, tailored to domains in which agent reputation is a key driver of the interactions between agents. The fundamentals are based on the principles of RepNet-POMDP, a framework developed by \textit{Rens et al.} \cite{rensetal} in 2018, but addresses its mathematical inconsistencies and alleviates its intractability by only considering fully observable environments. We furthermore use an online learning algorithm for finding approximate solutions to RepNet-MDPs. In a series of experiments, RepNet agents are shown to be able to adapt their own behavior to the past behavior and reliability of the remaining agents of the network. Finally, our work identifies a limitation of the framework in its current formulation that prevents its agents from learning in circumstances in which they are not a primary actor.

\end{abstract}

\section{Introduction}
Decision-making and learning in multi-agent settings is a multi-faceted area of research \cite{Boutilier,decmdp,decmdp2,moredec,ipomdp,nonstation,convergence}. Frameworks used for fully cooperative networks of agents differ vastly from those used for networks of self-interested agents.
A primary concern when dealing with self-centered agents is that it makes multi-agent learning inherently more complex than single-agent learning \cite{nonstation,convergence}. In fact, each agent needs to take into account the behavior of the entire network of agents when learning its own behavior. Additionally, agent behavior tends to be ever-changing. This \textit{non-stationarity} of agent behavior leads to the loss of policy \textit{convergence} properties that can often be found in single-agent formalisms \cite{convergence}.

In 2018, \textit{Rens et al.} \cite{rensetal} proposed a mathematical framework, called \textit{RepNet-POMDP}, designed to handle partially observable environments in which an agent's reputation among other agents dictates its behavior. The framework was subject to several mathematical inconsistencies, had no working implementation, and had a highly intractable planning algorithm.  

Nonetheless, the framework does present some ideas we believe are worth pursuing. Hence, in this paper, we provide an updated version of the framework, called RepNet-MDP. We address the mathematical inconsistencies of the original framework and alleviate its intractability by only considering fully observable environments. We furthermore make use of an online learning algorithm for finding approximate solutions to RepNet-MDPs. The viability of the framework is tested in a series of experiments designed to highlight its strengths and shortcomings. 

Section \ref{sec:background} summarizes the relevant background required. Section \ref{sec:related} gives an overview of the work related to our framework. Section \ref{sec:intuition} provides an intuitive introduction to RepNet-MDPs. Section \ref{sec:repnet} covers the formal definition of the framework. Section \ref{sec:planning} covers planning for RepNet-MDPs. The experimental setup and results are given in Section \ref{sec:exp}.

\section{Background - Markov Decision Processes}
\label{sec:background}

A Markov Decision Process (MDP) describes a process for modeling decision-making in stochastic environments \cite{Russell:2009:AIM:1671238}. An agent is assumed to move about in an environment, described by a set of states $\mathcal{S}$, by applying actions in $\mathcal{A}$ to the environment. The transition rules of the environment are dictated by the transition model $\mathcal{T} : \mathcal{S} \times \mathcal{A} \times \mathcal{S} \rightarrow [0,1]$, that is, $\mathcal{T}(s,a,s')$ returns the probability of the agent transitioning to state $s'$ upon performing action $a$ in state $s$. Each action applied to the environment results in a reward for the agent, dictated by the reward function $\mathcal{R}: \mathcal{S} \times \mathcal{A} \rightarrow \R$, that is, $\mathcal{R}(s,a)$ returns the reward received by the agent when performing action $a$ in state $s$.

The objective of an MDP agent is to maximize its long-term cumulative reward, called \textit{utility}. The utility $U$ of a \textit{finite} state-action sequence, sometimes called \textit{episode}, $E = \big \langle s_0,a_0,s_1, a_1,..., s_T, a_T \big \rangle$ is defined as \cite{Nitti2017}:
\begin{align*}
    U(E) = \sum_{t=0}^{T} \gamma^t \mathcal{R}(s_t, a_t),
\end{align*}
where $\gamma \in [0,1]$ is called the \textit{discount factor}. An agent advances in the environment by following a \textit{policy} $\pi : \mathcal{S} \times \N \rightarrow \mathcal{A}$ that maps each environment state and remaining time-steps to the action the agent should take.

The \textit{expected utility}, or \textit{value}, of being in any state $s_t$ at time-step $t$, while following policy $\pi$, with $d$ time-steps remaining, is defined as:
\begin{align*}
    V^{\pi}(s_t, d) = \E[U(E_t) \, | \, s_t, \pi] = \E \big[ \sum_{k = t }^{t+d} \gamma^{k-t} \mathcal{R}(s_k, a_k) \, \big| \, s_t, \pi \big ],
\end{align*}
where $E_t$ is the sub-sequence of $E$ starting at time-step $t$. An optimal policy $\pi^\star$ is a policy such that
\begin{align*}
    \forall s \in \mathcal{S}, \forall d \in \N, \forall \pi : V^{\star}(s,d) \geq V^{\pi}(s,d), 
\end{align*}
where $V^{\star} : \mathcal{S} \times \N \rightarrow \R$ is the value function associated with optimal policy $\pi^{\star}$.
This policy satisfies the \textit{optimality equations}, also known as the \textit{Bellman equations} ($\forall s \in \mathcal{S}$):
\begin{align*}
        \begin{dcases}
            V^{\star}(s, d) := \max_{a \in \mathcal{A}} \Big\{ \mathcal{R}(s,a) + \gamma \sum_{s' \in \mathcal{S}} \mathcal{T}(s,a,s') V^{\star}(s', d-1) \Big\}  &d > 1
            \\
            V^{\star}(s, 1) := \max_{a \in \mathcal{A}} \Big\{ \mathcal{R}(s,a) \Big\} &
        \end{dcases}
    \end{align*}

Partially Observable Markov Decision Processes are a common extension of classic MDPs that deal with the problem of partial observability of the environment \cite{Russell:2009:AIM:1671238}. To address the agent's inability to observe the exact state of the environment, the observation function $\mathcal{O} : \mathcal{A} \times \mathcal{S} \times \Omega \rightarrow [0,1]$, where $\Omega$ is the set of observations, is introduced. $\mathcal{O}(a,s',o)$ returns the probability of the agent making observation $o$ after performing action $a$ and the environment transitioning to state $s'$.

Instead of working with the actual states of the environment, the POMDP agents make use of the notion of belief state $b \in \Delta(\mathcal{S})$ \footnote{$\Delta (\mathcal{E})$ is the set of probability distributions over the elements of set $\mathcal{E}$.}, which is a probability distribution over the possible states of the environment. 
As such, $b(s)$ returns
the probability of being in state $s$. Furthermore,
\begin{align*}
    \sum_{s \in \mathcal{S}} b(s) = 1.
\end{align*}

Suppose the agent makes observation $o$ after taking action $a$ in current belief state $b$. The updated belief state $b'$ is computed using the \textit{state estimation function} $SE$ defined as follows:
\begin{align*}
    b' := SE(b,a,o) := \Big\{(s',p) \,\,\Big| \,\,s' \in \mathcal{S} \land p =  \frac{\mathcal{O}(a,s',o) \sum_{s} \mathcal{T}(s,a,s') b(s)}
    {P(o | b, a)} \Big\},
\end{align*}
where $P(o | b,a) = \sum_{s' \in \mathcal{S}} \mathcal{O}(a,s',o) \sum_{s \in \mathcal{S}} \mathcal{T}(s,a,s') b(s)$ is a normalizing constant. 
The optimal value function $V^{\star} : \Delta(\mathcal{S}) \times \N \rightarrow \R$ satisfies the following \textit{optimality equations} ($\forall b \in \Delta(\mathcal{S})$):

\begin{align*}
        \begin{dcases}
            V^{\star}(b, d) := \max_{a \in \mathcal{A}} \Big\{ \sum_{s \in \mathcal{S}}\mathcal{R}(s,a) b(s) + \gamma \sum_{o \in \Omega} P(o | b,a) V^{\star}(SE(b,a,o), d-1) \Big\}  
            \\
            V^{\star}(b, 1) := \max_{a \in \mathcal{A}} \Big\{ \sum_{s \in \mathcal{S}}\mathcal{R}(s,a) b(s) \Big\} 
        \end{dcases}
    \end{align*}
We refer to \cite{Russell:2009:AIM:1671238} for an extensive overview of POMDPs.

\section{Related MDP-based frameworks}
\label{sec:related}

Early multi-agent frameworks, such as Multi-agent Markov Decision Processes (MMDPs)  \cite{Boutilier} and Decentralized Partially Observable MDPs (Dec-POMDPs) \cite{decmdp,decmdp2}, operate under the assumption that the agents are selfless and have a common goal. Consequently, planning can be centralized, that is, each agent's policy can be computed by central unit, before being distributed amid the agents for execution \cite{moredec}. Dec-POMDPs furthermore differ from MMDPs in that states are no longer fully observable, meaning that each agent is in possession of its own set of \textit{local observations}.

In 2005, \textit{Gmytrasiewicz et al.} formalized an extension of POMDPs to multi-agent settings, called Interactive-POMDP (I-POMDP) \cite{ipomdp}. I-POMDPs are designed for reasoning in networks of selfish agents.
I-POMDP agents update their beliefs not only over physical states of the environment but also over models of the other agents in the network. The difficulty of solving I-POMDPs lies in the recursive nature of the models. Consider agent $g$'s belief update in a network inhabited by another agent, say $h$. A model of agent $h$ may consist of the belief function of said agent $h$ over physical states and models of all other agents. These models may, in turn, consist of belief functions of their own. This nesting of beliefs could theoretically be infinite, but is overcome by bounding the nesting depth by a finite number $n$, and solving the problem as a set of POMDPs.

The RepNet-MDP framework \cite{rensetal} simplifies the notion of model by focusing in on key concepts such as behavioral habits and reputation of other agents. While this reduces the insights RepNet agents can have into other agents' behavior, it makes the framework arguably more intuitive. The key, novel notion in the RepNet framework is that of subjective transitions, which have a dependence on the reputation of the agent performing the action.

\section{Developing an intuition for RepNet-MDPs}
\label{sec:intuition}

\begin{figure}[h]
\centering
\subfloat[The POMDP loop]
{
\label{fig:pomdp}
    \includegraphics[width=0.45\textwidth]{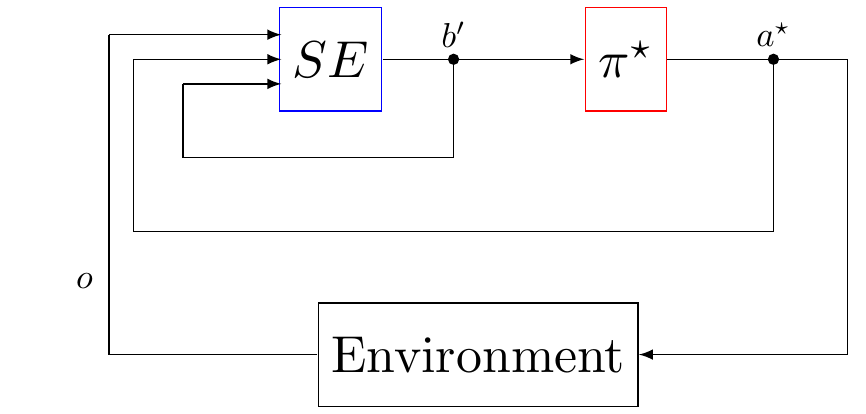}
}
\subfloat[The RepNet-MDP loop]
{
\label{fig:rep}
    \includegraphics[width=0.45\textwidth]{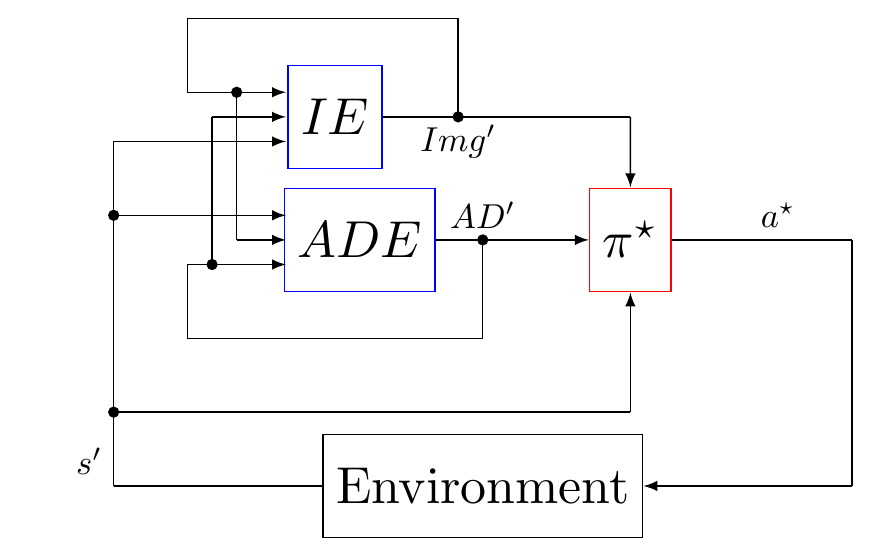}
}
\caption{POMDP and RepNet-MDP loops.}
\end{figure}

To develop an intuition for the RepNet-MDP framework, parallels between the concepts found in classic POMDPs and RepNet-MDPs can be drawn. In a POMDP, a single agent, placed in a partially observable environment, applies an action $a^{\star}$ it deems optimal as per its current policy $\pi^{\star}$, and is sent back an observation $o$. The state estimation function $SE$ can be thought of as a way of extracting information from said observation $o$, and storing it in a belief state $b'$. More specifically, $o$ contains information about the actual state of the environment. The POMDP loop is depicted in Fig. \ref{fig:pomdp}.

Let us now consider a fully observable environment made up of 3 selfish agents, of which the behavior of the first is dictated by the RepNet-MDP framework. The willingness of the RepNet agent to engage with agents 2 or 3 is to be conditioned by their reputation and behavioral habits. The first agent once again applies action $a^{\star}$, as per its policy $\pi^{\star}$. The environment returns its new state $s'$. In an effort to make well-informed decisions, the RepNet agent should extract the other agents' behavior from $s'$. 

Two functions, analogous to the state estimation function $SE$ in POMDPs, are used to this end:
 The action distribution estimation function $ADE$ extracts information regarding other agents' behavioral habits.
 The image estimation function $IE$ informs the RepNet agent on the image all the agents have of each other.
The RepNet-MDP loop is shown in Fig. \ref{fig:rep}.

Closely tied to the concept of image is the notion of reputation. Specifically, the reputation of any agent in the framework can be seen as a summary of the information encapsulated by the image.
Unlike POMDPs, RepNet-MDPs feature two types of actions, and by extention two types of transition models:

\begin{itemize}
    \item \textit{Objective} actions, which, when performed, have a real effect on the environment. These actions can be seen as equivalent to actions as they exist in MDPs.  The associated transition model is called the objective transition model $OT$ and describes the rules of the environment as they apply to the RepNet agent.
    \item \textit{Subjective} actions, which, unlike objective actions, are never actually applied to the environment. Instead, they are associated with another transition model called the subjective transition model $ST$: This transition model describes a RepNet agent's subjective perception of the rules of the environment. This perception is a function of said agent's reputation, and can be used by the agent to aid in its decision-making.
\end{itemize}

\section{Formal definition of RepNet-MDPs}
\label{sec:repnet}
In this section, we will formalize the RepNet-MDP framework introduced in Section \ref{sec:intuition}. A RepNet-MDP $\mathcal{M}$ is defined as a pair of tuples

\begin{align*}
    \mathcal{M} := \big \langle \Sigma, \Gamma \big \rangle,
  \end{align*}
where $\Sigma$ is called the System tuple and incorporates aspects of the network that apply to all agents, and $\Gamma$ is called the Agents tuple and contains each agent's subjective understanding of the environment it operates in.

Specifically, a System in a RepNet-MDP $\Sigma$ is formally defined as a tuple
\begin{align*}
    \Sigma := \big \langle \mathcal{G}, \mathcal{S}, \mathcal{A}, \mathcal{I}, \mathcal{U}, OT \big \rangle,
\end{align*}

where:
\begin{itemize}
    \item $\mathcal{G}$ is the set of agents that can interact with the environment.
    \item $\mathcal{S}$ is the set of possible states of the environment.
    \item $\mathcal{A}$ is the set of possible actions, both \textit{objective} and \textit{subjective}. Formally,
    \begin{align*}
    \mathcal{A} := \mathcal{A}^o \cup \mathcal{A}^s\,\,\,\,\,\,\,\,\,\,\,
    \mathcal{A}^o \cap \mathcal{A}^s := \varnothing.
  \end{align*}
The concept of \textit{subjective actions} will be discussed in Section \ref{sec:dirr}.
  
  \item $\mathcal{I}: \mathcal{G} \times \mathcal{G} \times \mathcal{S} \times \mathcal{A} \rightarrow [-1,1]$ is called the \textit{impact function}. $\mathcal{I}(g,h,s,a)$ returns the impact on agent $g$ that is due to agent $h$ performing action $a$ in state $s$. This function can be thought of as analogous to a Markov Decision Process's immediate reward function $\mathcal{R}$.
  \item $\mathcal{U} : [-1,1] \times [-1,1] \rightarrow [-1,1]$ is called the \textit{image update function}. Given a current value $v$, of the image an agent has of another agent, to be updated, and a new expected total impact $i$, of which the definition will be given shortly, $\mathcal{U}(v, i)$ returns an updated value of the image $v'$.
  Many instantiations of this function are possible, two of which are presented in \cite{rensetal}. We will use the following instantiation:
    \begin{align*}
    \mathcal{U}(v, i) :=
        \begin{dcases}
            v + (1-v)i &\text{ if } i \geq 0
            \\
            v + (1+v)i &\text{ if } i < 0
        \end{dcases}
    \end{align*}
\item $OT : \mathcal{G} \times \mathcal{S} \times \mathcal{A}^o \times \mathcal{S} \rightarrow [0,1]$ is called the \textit{objective transition model}. $OT(h,s,a,s')$ returns the probability of the environment transitioning from state $s$ to state $s'$ when objective action $a$ is taken by agent $h$.
\end{itemize}

In addition to the global information stored in $\Sigma$, each RepNet agent's subjective knowledge is stored in the Agents tuple $\Gamma$, formally defined as \footnote{$\{X_g\}$ is used as the shorthand notation for $\{X_g \,\,\lvert\,\, g \in \mathcal{G}\}$}
\begin{align*}
    \Gamma := \big \langle \{ST_g\}, \{AD_g\}, \{Img_g\} \big \rangle,
  \end{align*}

where:
\begin{itemize}
    \item $ST_g : \mathcal{G} \times \mathcal{S} \times \mathcal{A}^s \times \mathcal{S} \times [-1,1] \rightarrow [0,1]$ is called the \textit{subjective transition model} of agent $g$. $ST_g(h,s,a,r_h,s')$ returns the probability, \textit{as perceived by agent $g$}, of the environment transitioning from state $s$ to state $s'$ if agent $h$ were to perform subjective action $a$, and has a reputation $r_h$ according to agent $g$.
    
    \item $AD_g : \mathcal{G} \times \mathcal{S} \rightarrow \Delta(\mathcal{A})$ is called the \textit{action distribution} according to agent $g$. $AD_g(h,s)$ returns a probability distribution over actions in $\mathcal{A}$ for agent $h$ in state $s$, according to agent $g$.
    \item $Img_g : \mathcal{G} \times \mathcal{G} \rightarrow [-1,1]$ is called the \textit{image function} according to agent $g$. $Img_g(h,i)$ returns the image agent $i$ has of agent $h$ according to agent $g$. Said differently, it returns what $g$ thinks $i$ thinks of $h$.

\end{itemize}

As introduced in Section \ref{sec:intuition}, every agent bases its decision-making on the image it believes all agents to have of each other, as well as each agent's behavioral habits. Let $g$ be an agent, whose image at time $t$ is $Img_g$, and action distribution is $AD_g$. At time $t+1$, these constructs are updated via the image estimation function $IE$ and action distribution estimation function $ADE$ respectively, to produce $Img_g'$ and $AD_g'$.

\subsection{Image and Reputation}
This subsection builds towards the formal definition of the image estimation function $IE$. To this end, we introduce the notion of \textit{expected total impact}. Consider two agents $h$ and $i$. In any given state, agent $h$ can perform one of several actions which may or may not have an impact on agent $i$. Likewise, agent $i$ can be expected to have an impact on agent $h$ when performing an action. The \textit{expected total impact} should be thought of as a way of assigning a numerical value to the \textit{bidirectional} impact these two agents can be expected to have on each other. Additionally, one direction of the impact may be perceived as more important than the other and thus be weighed differently.
According to an observing agent, say $g$, the total impact $h$ is expected to have on, as well as perceive from, $i$ when the environment is in state $s$, is defined as
\begin{align*}
\begin{split}
        ETI_{g} (h, i, s, AD_g) :=  \sum_{a \in \mathcal{A}} \, \big[ \, \delta AD_g(i, s)(a) \mathcal{I}(h, i, s, a)  \\+ (1 - \delta) AD_g(h, s)(a) \mathcal{I}(i, h, s, a) \, \big],
\end{split}
\end{align*}
where $\delta \in [0,1]$ weighs the importance of impact due to agent $h$ and impact perceived by $h$.

Agent $g$'s image of other agents, as well as the image it believes all agents to have of each other changes as it observes the agents' behavior. 
Let $Img_g$ be the current image function of agent $g$. 
Concretely, we wish to \textit{update} the image any agent $i$ has of any other agent $h$, according to the observing agent $g$ ($= Img_g(h,i)$) on the basis of the impact $h$ is expected to have on $i$ ($= ETI_{g} (h, i, s, AD_g)$).
The updated image function $Img_g'$ is computed as follows:
\begin{align*}
    \begin{split}
        Img_g' &:= IE(g,Img_g,\alpha, s, AD_g) \\&:= \Big\{(h,i, t) \,\,\Big| \,\,h,i \in \mathcal{G} \land t =  \mathcal{U}(Img_g(h,i), ETI_{g}(h, i,s, AD_g)) \Big\},
    \end{split}
    \end{align*}
where $s$ is the current state of the environment, $IE$ is called the \textit{image estimation function},
and $\mathcal{U}$ is the \textit{image update function}.

Finally, the notion of reputation as it is understood in this framework can be thought of as a way of summarizing the information encapsulated by the image. 

Say agent $g$ wishes to estimate the reputation of agent $h$ in a network made up of several other agents. It can, to this end, use the image each agent $i$ has of agent $h$ ($=Img_g(h,i)$) as a guiding principle. A first idea might be to take agent $h$'s reputation to be equal to its average image in the network. If, however, some agent $i$ has a poor image of agent $h$ ($Img_g(h,i) < 0$), but agent $g$ has a poor image of agent $i$ ($Img_g(i,g) < 0$), it may be unreasonable for agent $g$ to assume that agent $i$'s opinion of agent $h$ is indicative of agent $h$'s reputation being poor.
These concerns are addressed by weighing the image each agent $i$ has of $h$ by the image $g$ has of $i$. As such, if both images are negative, the resulting reputation of $h$ will not be affected negatively ($Img_g(h,i) \times Img_g(i,g) > 0$).

Formally, the reputation of an agent $h$, according to agent $g$, is defined as 
\begin{align*}
      REP_g(h, Img_g) := \frac{1}{|\mathcal{G}'|} \sum_{i \in \mathcal{G}'} Img_g(h,i) \times Img_g(i,g),
\end{align*}
where $Img_g(i,i) = 1 \,\,\forall i \in \mathcal{G}$, and $\mathcal{G}' = \mathcal{G}$ if $h \neq g$ and $\mathcal{G}' = \mathcal{G} \setminus \{ g \}$ if $h = g$. Recall that in the RepNet framework, reputation influences subjective transition probabilities, which, in turn, influence a RepNet agent's planning.

\subsection{Subjective actions and the subjective transition model}
\label{sec:dirr}
\begin{figure}
  \begin{center}
    \includegraphics[width=0.47\textwidth]{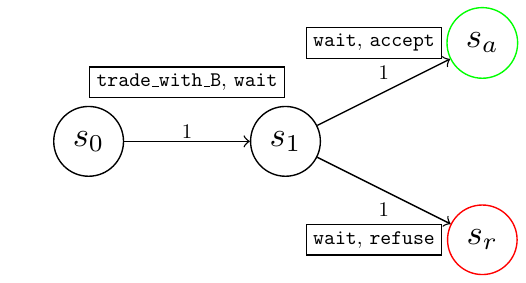}
  \end{center}
  \caption{Transition model of the environment (trading example). Each transition has two \textit{objective} actions, the first action represents agent $A$'s \textit{objective} action, the second action represents agent $B$'s \textit{objective} action.}\label{fig:dir2}
\end{figure}

In this section, we describe the use of subjective actions and subjective transition models in the RepNet framework.
As introduced in Section \ref{sec:intuition}, we make a distinction between the purpose of an \textit{objective} transition model, which describes the actual rules of the environment as they apply to the RepNet agent, and that of a \textit{subjective} transition model, which describes that agent's \textit{subjective} perception of the rules of the environment, this perception being influenced by the reputation of the RepNet agent. To illustrate this further, we will make use of a simple trading example between two agents $A$ and $B$. Agent $A$ wishes to trade with agent $B$, who can either accept or refuse the trade offer. The environment is made up of the set of states 
\begin{align*}
    \mathcal{S} = \{s_0, s_1, s_a, s_r\}.
\end{align*}
 $s_0$ is the initial state, prior to any trade, $s_1$ is the state in which agent $B$ is made aware of agent $A$'s trade offer, $s_a$ is the accept state, and $s_r$ is the refuse state. 
 The set of \textit{objective} actions at the disposal of both agents is given by
\begin{align*}
    \mathcal{A}^o = \{\texttt{trade\_with\_A}, \texttt{trade\_with\_B}, \texttt{accept}, \texttt{refuse}, \texttt{wait}\}.
\end{align*}
The transition model of the environment assumed to be deterministic, is given in Fig. \ref{fig:dir2}.
\textit{In the eyes of} agent $A$, agent $B$'s response to a trade offer, characterized by transitions $s_1 \rightarrow s_a$ and $s_1 \rightarrow s_r$, depends on $A$'s reputation. The action taken by agent $A$ during these transitions is \texttt{wait}. To make use of the notion of subjective actions, the set of \textit{subjective} actions $\mathcal{A}^s$ will contain the counterpart\footnote{We define \textit{counterpart} as a partial mapping $\mathcal{C} : \mathcal{A}^o \rightarrow \mathcal{A}^s$. If $\mathcal{C}(a)$ is not defined, then $a$ has no counterpart in $\mathcal{A}^s$. } of \texttt{wait} in its \textit{subjective} form, that is,
\begin{align*}
    \mathcal{A}^s = \{\texttt{wait\_s}\}.
\end{align*}

The way agent $A$ makes use of actions in $\mathcal{A}^o$ and $\mathcal{A}^s$ can now be detailed.
When \textit{planning} to maximize its expected impact, agent $A$ will make use of the \textit{objective} transition model whenever the action currently investigated has no subjective counterpart in $\mathcal{A}^s$. For instance, the transition probability when investigating action \texttt{trade\_with\_B} is given by
\begin{align*}
   OT(A, s_0, \texttt{trade\_with\_B}, s_1).  
\end{align*}
    When an action in $\mathcal{A}^o$ has a counterpart in $\mathcal{A}^s$, agent $A$ will make use of the subjective transition model. For instance, the transition probability when investigating action \texttt{wait}/\texttt{wait\_s} is given by
    \begin{align*}
       ST_A(A, s_1, \texttt{wait\_s}, s_a, r_A).
    \end{align*}

As such, the reputation of agent $A$ is accounted for when agent $A$ \textit{plans} to maximize its expected impact.

\subsection{Action distribution}
\label{sec:revv}

The next step in the formalization of RepNet-MDPs consists in redefining the updating scheme of the action distribution $AD_g$ of each agent $g$. Say an environment hosting two agents $g$ and $h$ is currently in state $s$. Agent $g$ has an \textit{a priori} notion of the probability of agent $h$ picking an action $a$ in state $s$,
\begin{align*}
    P_g(a | h, s,r_h).
\end{align*}
Following agent $h$ performing action $a$ in this state, the environment transitions from state $s$ to state $s'$. The \textit{a posteriori} probability of agent $h$ performing that same action $a$ in state $s$ in the future is now computed using Bayes' rule\footnote{Bayes' theorem is defined mathematically as follows: $P(A | B,C) = \frac{P(B|A,C) P(A|C)}{P(B|C)}$, where $A$, $B$ and $C$ are events and $P(B | C) \neq 0$.}:
\begin{align*}
\begin{split}
    P_g(a | h,s,r_h,s') &= \frac{P_g(s' | h,s,r_h,a) P_g(a | h, s,r_h)}{P_g(s'|h,s,r_h)} 
    \\
    &= \frac{P_g(s' | h,s,r_h,a) P_g(a | h, s)}{\sum_{a'} P_g(s'|h,s,r_h,a') P_g(a' | h,s)}.
\end{split}
\end{align*}
The probabilities may now be replaced by the RepNet nomenclature:
\begin{align*}
    AD_g'(h,s)(a) = \frac{T_g(h,s,a,s', r_h) AD_g(h,s)(a)}{\sum_{a'} T_g(h,s,a',s', r_h) AD_g(h,s)(a')}.
\end{align*}

One can add   \textit{smoothing}
smoothing to the present result in an effort to
avoid undesirable side effects of using a deterministic transition model.
Consider the trading scenario between agents A and B described in Section
\ref{sec:dirr}. The transitions of the environment are assumed to be deterministic, that is
\begin{align*}
    UT(B, s_1, \texttt{accept}, a) = 1, \,\,\,\,\, UT(B, s_1, \texttt{refuse}, r) = 1.
\end{align*}
If agent $B$ refuses the trade offer made by agent $A$, the environment transitions to state $r$ and the action distribution is updated as follows:

\begin{align*}
    AD_A'(B,s_1)(\texttt{accept}) &=   \frac{UT(B,s_1, \texttt{accept},r) AD_A(B,s_1)(\texttt{accept})}{\sum_{a'} UT(B,s_1,a',r) AD_A(B,s_1)(a')}
    \\&=  \frac{0 \cdot AD_A(B,s_1)(\texttt{accept})}{\sum_{a'} UT(B,s_1,a',r) AD_A(B,s_1)(a')}
    \\&= 0
\end{align*}

\begin{align*}
    AD_A'(B,s_1)(\texttt{refuse}) &=   \frac{UT(B,s_1, \texttt{refuse},r) AD_A(B,s_1)(\texttt{refuse})}{\sum_{a'} UT(B,s_1,a',r) AD_A(B,s_1)(a')}
    \\&=  \frac{1 \cdot AD_A(B,s_1)(\texttt{refuse})}{\sum_{a'} UT(B,s_1,a',r) AD_A(B,s_1)(a')}
    \\&= 1
\end{align*}

As such, agent A is, in the wake of a single unsuccessful trade, now convinced
that agent B will never accept any trade offer in the future. Moreover, it is now
impossible for agent A to change its strategy in the future. In fact, the probability
of B accepting a trade is 0, and regardless of what this value is multiplied by in
the future, it will always remain 0. 

This inconvenience is addressed by applying a \textit{smoothing} technique called \textit{Laplace smoothing} \cite{laplacesmoothing}. The smoothing technique prevents probabilities of 0 from ever occurring, and can be applied to the \textit{action distribution update} function, resulting in the following definition for the \textit{action distribution estimation}:

\begin{align*}
\begin{split}
     AD_g' &:= ADE(g, s', AD_g, Img_g) \\&:= \Bigg\{(h,s,a,p) \,\,\Bigg| \,\, h \in \mathcal{G} \land s \in \mathcal{S} \land a \in \mathcal{A} \,\land r_h = REP_g(h,Img_g)
     \\& \,\,\,\,\,\,\,\,\,\,\, \land p =  \frac{T_g(h,s,a,s', r_h) AD_g(h,s)(a) + \eta}{\sum_{a'} (T_g(h,s,a',s', r_h) AD_g(h,s)(a') + \eta)} \Bigg\},
\end{split}
\end{align*}
where $ADE$ is called the \textit{action distribution estimation function}, $s'$ is the state the environment transitions to, and $\eta$ is the \textit{Laplace smoothing parameter}.

Note that to simplify the notation, we combined the \textit{objective} and \textit{subjective} transition models into a single model $T_g$, called the \textit{global transition model} and formally defined as
\begin{align}
\label{eq:tg}
     T_g(h, s,a_h,s', r_h) := 
    \begin{cases}
    ST_g(h, s,a_h,s', r_h) & \mbox{if } a_h \in \mathcal{A}^s
    \\
    OT(h,s,a_h,s') & \mbox{if } a_h \in \mathcal{A}^o
    \end{cases}
\end{align}

\section{Planning in the RepNet framework}
\label{sec:planning}

We now describe optimal behavior in the context of RepNet-MDPs, for finite horizon look-ahead. To simplify the notation, we can define a construct called \textit{epistemic state}.
The epistemic state $\theta_g$ of agent $g$ is formally defined as a tuple
\begin{align*}
     \theta_g := \big \langle s,  AD_g, Img_g \big \rangle,
\end{align*}
where $s$ is the current state of the environment,
    $AD_g$ is the current action distribution of agent $g$,
and $Img_g$ is the current image function of agent $g$.
$\theta_g \in \Theta_g$, and $\Theta_g$ is called the \textit{epistemic state space}. This set contains every possible combination of physical states of the environment, action distributions, and image functions of agent $g$.

An agent should perform actions according to the \textit{perceived immediate impact} they have on the agent itself. 
The perceived immediate impact on agent $g$ resulting from performing action $a$ in state $s$ is defined as
    \begin{align*}
        PI_g(s,AD_g, a) := \frac{1}{|\mathcal{G}|} \, \big[ \, \mathcal{I}(g,g,s,a) + \sum_{h \in \mathcal{G} \setminus \{ g \}}  \sum_{a' \in \mathcal{A}} \mathcal{I}(g, h, s, a') AD_g(h, s)(a') \, \big],
    \end{align*}
    where
$AD_g$ is the current action distribution of agent $g$.
The first term describes the immediate self-impact as a consequence of agent $g$ performing action $a$, while the second term describes the expected immediate impact that the network (i.e., the remaining agents) has on agent $g$.

Analogously to regular MDPs, a RepNet-MDP agent $g$ strives to maximize its expected discounted perceived impact
\begin{align*}
    \E \big[ \sum_{t = 0}^{k} \gamma^t PI_{g,t}\big],
\end{align*}
where $\gamma$ is the \textit{discount factor} and $PI_{g,t}$ is agent $g$'s perceived immediate impact at time-step $t$. This is accomplished by computing the optimal value function $V_g : \Theta_g \times \N \rightarrow \R$ (in a \textit{finite-horizon} setting). It satisfies the \textit{optimality equations}, which are defined as ($\forall \theta_g \in \Theta_g$)
    \begin{align}
        \begin{dcases}
            V_g(\theta_g, k) := \max_{a \in \mathcal{A}} \Big\{ PI_g(s,AD_g,a) + \gamma \sum_{s' \in \mathcal{S}} T_g(g,s,a,s',r_g) V_g( \theta_g', k-1) \Big\}  
            \\
            V_g(\theta_g, 1) := \max_{a \in \mathcal{A}} \Big\{ PI_g(s,AD_g,a) \Big\} 
        \end{dcases}
        \label{eq:een}
    \end{align}
    where 
 $r_g = REP_g(g, Img_g)$, $\theta_g = \big \langle s, AD_g, Img_g \big \rangle$, and \\ $\theta_g' = \big \langle s', ADE(g, s', AD_g, Img_g), IE(g, Img_g, \alpha, s, AD_g) \big \rangle$.

The optimal action for agent $g$, written $\pi(\theta_g, k)$, is defined as
\begin{align*}
    \pi(\theta_g, k) := \arg \max_{a \in \mathcal{A}} \Big\{ PI_g(s,AD_g,a) + \gamma \sum_{s' \in S} T_g(g,s,a,s', r_g) V_g( \theta_g', k-1) \Big\}.
\end{align*}

In this work, we implement (approximate) online planning \cite{offlineonline} instead of exact planning.
The general principle of model-based online planning can be described as the interleaving of two phases, the \textit{planning phase}, in which the (PO)MDP performs a look-ahead search of a given depth $D$, starting at the current environment state, the goal being to determine the most suitable action, and the \textit{execution phase}, in which this action is applied to the environment \cite{offlineonline}. The description of online planning applied to the RepNet-MDP framework follows hereafter. 

Let $g$ be an agent deployed in an environment that can be in two physical states $s_0$ or $s_1$. The set of possible actions $\mathcal{A}$ comprises actions $a_0$ and $a_1$, and the environment is currently in state $s_0$. Agent $g$ has current action distribution $AD_g^0$ and image function $Img_g^0$. The physical state of the environment, action distribution, and image function can be combined to form an \textit{epistemic state} $\theta_g^0 = \langle s_0, AD_g^0, Img_g^0 \rangle$.

\begin{figure}[h]
  \begin{center}
    \includegraphics{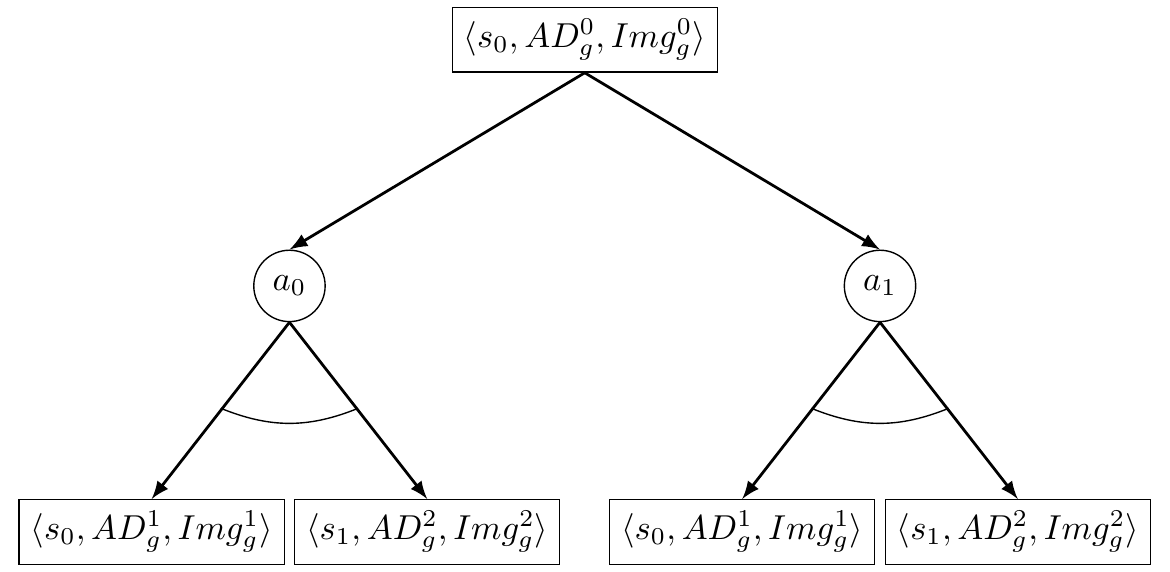}
  \end{center}
  \caption{Look-ahead search space, depth $D=1$ (RepNet-MDP)}\label{fig:repnettree}
\end{figure}

The RepNet agent can construct a look-ahead tree of depth $D$ starting at the current \textit{epistemic state}. Fig. \ref{fig:repnettree} depicts a search space of depth $D = 1$. Every action in $\mathcal{A}$ first leads to the formation of a new branch, at the end of which is an AND-node (or \textit{action} node) with the corresponding action. From each AND-node, every physical state is a potential future state and requires the formation of a new branch at the end of which is an OR-node (or \textit{epistemic state} node) that contains the corresponding \textit{epistemic state}. Consider the \textit{epistemic state} in the lower-left corner of Fig. \ref{fig:repnettree}, i.e. $\langle s_0, AD_g^1, Img_g^1 \rangle$. $AD_g^1$ and $Img_g^1$ are obtained by updating the previous action distribution $AD_g^0$ and the previous image function $Img_g^0$, using the new physical state $s_0$. 

After constructing the search space, an estimation of the value function needs to be back-propagated from the leaves to the root of the tree. A heuristic estimate of the true value function
\begin{align*}
    h : \Theta_g \rightarrow \R
\end{align*}
can be computed at the leaves by taking the base case of the Bellman equations for RepNet-MDPs:
\begin{align*}
    h(\theta_g) = \max_{a \in \mathcal{A}} \Big\{ PI_g(s,AD_g,a) \Big\}.
\end{align*}
The \textit{epistemic state-action} values are then computed at the action nodes as follows:
\begin{align*}
     q(\theta_g, a) = PI_g(s,AD_g,a) + \gamma \sum_{s' \in \mathcal{S}} T_g(g,s,a,s',r_g) h(\theta'_g).
\end{align*}
Non-leaf state nodes use the maximum of the \textit{epistemic state-action} values of their children as their estimate of the value function, i.e.
\begin{align*}
    h(\theta_g) = \max_{a \in \mathcal{A}} \Big\{ q(\theta_g, a)  \Big\}.
\end{align*}

\section{Experiments}
\label{sec:exp}

The goal of the experiments is to showcase the strengths and shortcomings of the framework. To this end, the experimental setup consists of 2 trading scenarios, for which several experiments are conducted. All experiments were conducted with \textit{look-ahead depth} $D = 3$, \textit{Laplace smoothing parameter} $\eta = 0.1$, and \textit{discount factor} $\gamma = 0.7$.
Note that these experiments serve as a proof of concept for the RepNet framework and, as such, are not designed to reflect the framework's applicability to problems of realistic scale.

\subsection{Experiment 1: Trading between two agents} \label{ex:trade2ag}
Let $A$ and $B$ be two agents. Agent $A$ plays the role of the buyer, agent $B$ the role of the seller. Agent $A$ can engage in a trade with agent $B$, and $B$ can accept or refuse the trade offer. Furthermore, agent $A$ can, prior to making a trade offer, do a good deed in an effort to improve its image in the eyes of agent $B$. 

In this series of experiments, Agent $A$ is managed by the RepNet algorithm. Agent $B$ is run by a simple algorithm that accepts or rejects trade offers made by agent $A$ according to a set schedule. In particular, agent $B$ is asked to reject trade offers for the 20 first time-steps, accept them for the 60 subsequent time-steps, and finally reject them for the last 20 time-steps.

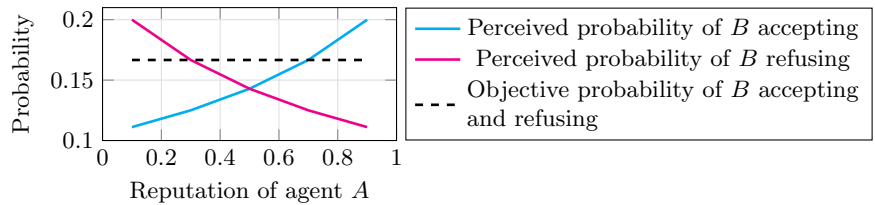
\begin{figure}[h]
\centering

\begin{tikzpicture}
\begin{axis}[
    xlabel={Reputation of agent $A$},
	ylabel= Probability,
	title style={align=left},
	grid=both,
	minor grid style={gray!25},
	major grid style={gray!25},
	legend pos=outer north east,
	width=0.45\linewidth,
	height=3.35cm,
	legend style={cells={align=left}},
	no marks,
	xmin=0,
    xmax=1,
    ymin=0.1,
    ymax=0.21
    ]

\addplot+[line width=1pt,solid,color=cyan] plot coordinates { (0.1, 1/9) (0.3, 1/8) (0.5, 1/7) (0.7, 1/6) (0.9, 1/5)};
\addlegendentry{Perceived probability of $B$ accepting};

\addplot+[line width=1pt,solid,color=magenta] plot coordinates { (0.1, 1/5) (0.3, 1/6) (0.5, 1/7) (0.7, 1/8) (0.9, 1/9)};
\addlegendentry{Perceived probability of $B$ refusing};

\addplot+[line width=1pt,dashed,color=black] plot coordinates { (0.1, 1/6) (0.3, 1/6) (0.5, 1/6) (0.7, 1/6) (0.9, 1/6)};
\addlegendentry{Objective probability of $B$ accepting \\and refusing}; 

\end{axis}
\end{tikzpicture}

\caption{Perceived probability of agent $B$ accepting and refusing the trade offers, as a function of the self-reputation of agent $A$.}
\label{figprofile}
\end{figure}

Two series of experiments are conducted, the first one without making use of subjective actions, the second one by modeling the  action of agent $A$ awaiting agent $B$'s response to a trade offer as a subjective action, meaning the outcome of agent $A$'s planning will be influenced by its reputation. A well-designed subjective transition model, schematized in Fig. \ref{figprofile}, that realistically reflects how the reputation of agent $A$ may influence the willingness of agent $B$ to accept $A$'s trade offers is put to the test.
The variables tracked are the action distribution, image, and by extension the reputation of both agents \textit{in the eyes of agent $A$}, and frequency at which agent $A$ makes trade offers. 

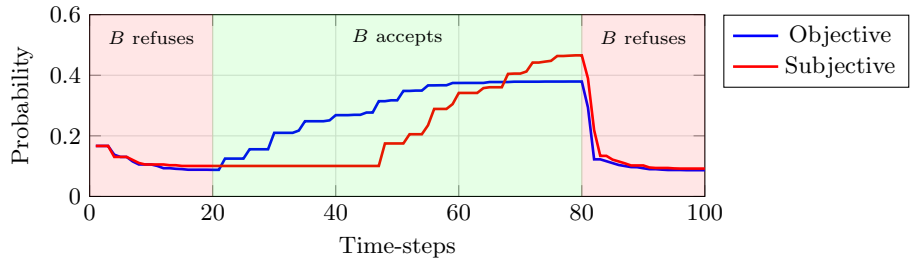
\begin{figure}
\centering

\begin{tikzpicture}
\begin{axis}[
	xlabel=Time-steps,
	ylabel= Probability,
	title style={align=left},
	grid=both,
	minor grid style={gray!25},
	major grid style={gray!25},
	legend pos=outer north east,
	width=0.8\linewidth,
	height=4cm,
	no marks,
	xmin=0,
    xmax=100,
    ymin=0,
    ymax=0.6]
\addplot[line width=1pt,solid,color=blue] %
	table[x=b,y=a,col sep=comma]{tradeABaccept2u.csv};
\addlegendentry{Objective};

\addplot[line width=1pt,solid,color=red] %
	table[x=b,y=a,col sep=comma]{tradeABaccept2.csv};
\addlegendentry{Subjective};


\addplot+[draw=red,fill=red, opacity=0.1]
        coordinates {(0,0) (0,2) (20,2) (20,0)};

\addplot+[draw=green,fill=green, opacity=0.1]
        coordinates {(20,0) (20,2) (80,2) (80,0)};
        
\addplot+[draw=red,fill=red, opacity=0.1]
        coordinates {(80,0) (80,2) (100,2) (100,0)};

\node[] at (10,60pt) {\scriptsize{$B$ refuses}};    
\node[] at (50,60pt) {\scriptsize{$B$ accepts}};    
\node[] at (90,60pt) {\scriptsize{$B$ refuses}}; 

\end{axis}
\end{tikzpicture}

\caption{Probability of $B$ accepting $A$'s trade offers, according to $A$}
\label{fig1}

\end{figure}

\begin{figure}
\centering

\begin{tikzpicture}
\begin{axis}[
	xlabel=Time-steps,
	ylabel= Reputation,
	title style={align=left},
	grid=both,
	minor grid style={gray!25},
	major grid style={gray!25},
	legend pos=outer north east,
	width=0.8\linewidth,
	height=4cm,
	no marks,
	xmin=0,
    xmax=100,
    ymin=-0.3,
    ymax=1]

\addplot[line width=1pt,solid,color=red] %
	table[x=b,y=a,col sep=comma]{tradeABrepA.csv};
\addlegendentry{Subjective};


\addplot+[draw=red,fill=red, opacity=0.1]
        coordinates {(0,-1) (0,2) (20,2) (20,-1)};

\addplot+[draw=green,fill=green, opacity=0.1]
        coordinates {(20,-1) (20,2) (80,2) (80,-1)};
        
\addplot+[draw=red,fill=red, opacity=0.1]
        coordinates {(80,-1) (80,2) (100,2) (100,-1)};
    
\node[] at (10,43pt) {\scriptsize{$B$ refuses}};    
\node[] at (50,43pt) {\scriptsize{$B$ accepts}};    
\node[] at (90,43pt) {\scriptsize{$B$ refuses}}; 
\end{axis}
\end{tikzpicture}

\caption{Reputation of agent $A$, according to itself.}
\label{fig2}
\end{figure}
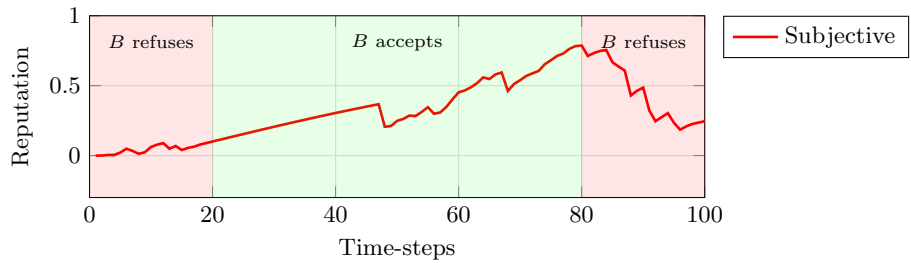

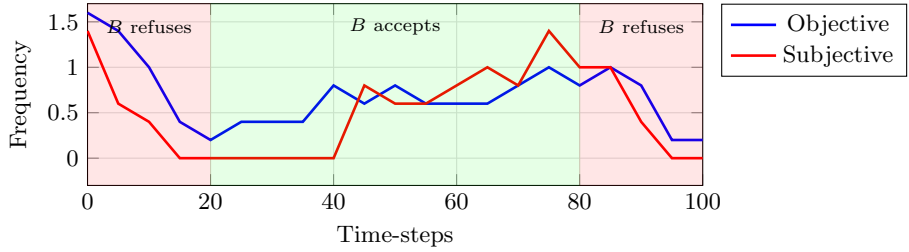
\begin{figure}
\centering

\begin{tikzpicture}
\begin{axis}[
    xlabel=Time-steps,
	ylabel= Frequency,
	title style={align=left},
	grid=both,
	minor grid style={gray!25},
	major grid style={gray!25},
	legend pos=outer north east,
	width=0.8\linewidth,
	height=4cm,
	no marks,
	xmin=0,
    xmax=100,
    ymin=-0.3,
    ymax=1.7
    ]
\addplot+[line width=1pt,solid,color=blue] plot coordinates { (0, 1.6) (5, 1.4) (10, 1) (15, 0.4) (20, 0.2) (25, 0.4)(30, 0.4)(35, 0.4)(40, 0.8)(45, 0.6)(50, 0.8)(55, 0.6)(60, 0.6)(65, 0.6)(70, 0.8)(75, 1)(80, 0.8)(85, 1)(90, 0.8)(95, 0.2)(100, 0.2)};
\addlegendentry{Objective};

\addplot+[line width=1pt,solid,color=red] plot coordinates { (0, 1.4) (5, 0.6) (10, 0.4) (15, 0) (20, 0) (25, 0)(30, 0)(35, 0)(40, 0)(45, 0.8)(50, 0.6)(55, 0.6)(60, 0.8)(65, 1)(70, 0.8)(75, 1.4)(80, 1)(85, 1)(90, 0.4)(95, 0)(100, 0)};
\addlegendentry{Subjective};


\addplot+[draw=red,fill=red, opacity=0.1]
        coordinates {(0,-1) (0,2) (20,2) (20,-1)};

\addplot+[draw=green,fill=green, opacity=0.1]
        coordinates {(20,-1) (20,2) (80,2) (80,-1)};
        
\addplot+[draw=red,fill=red, opacity=0.1]
        coordinates {(80,-1) (80,2) (100,2) (100,-1)};

\node[] at (10,50pt) {\scriptsize{$B$ refuses}};    
\node[] at (50,50pt) {\scriptsize{$B$ accepts}};    
\node[] at (90,50pt) {\scriptsize{$B$ refuses}}; 

\end{axis}
\end{tikzpicture}

\caption{Frequency of the trade offers made by $A$, measured in 5 time-step intervals}
\label{fig3}
\end{figure}

Fig. \ref{fig1} shows the evolution of agent $A$'s action distribution for target agent $B$.
Fig. \ref{fig2} shows the evolution of agent $A$'s self-reputation during the experiment involving the subjective transition model. Note that $A$'s self-reputation, and more generally $A$'s image function, have no bearing on its decision-making if no subjective actions are used (see Equations \ref{eq:tg} and \ref{eq:een}, $T_g$ makes use of the notion of reputation only for subjective actions).
Finally, Fig. \ref{fig3} shows the evolution of the frequency at which $A$ makes trade offers.

In the first 20 time-steps, $B$ refuses each trade offer. Regardless of the series of experiments, agent $A$ is able to pick up on this via the action distribution. As a consequence, it quickly reduces the frequency at which it attempts to trade with $B$. 
In the 60 following time-steps, $B$ is asked to change its behavior and accept each trade offer. Hesitant at first, $A$ gradually increases the frequency at which it attempts to trade with $B$.
Agent $A$ is able to pick up on $B$ reverting back to its old behavior during the final 20 steps.

Additionally making use of a well-designed subjective transition model noticeably improves the RepNet agent's performance. While the trajectories showcase the same key elements, the pace at which agent $A$ is able to adapt improves greatly.
The subjective transition model was designed such that agent $A$ believes that its reputation must be good for $B$ to be willing to trade with $A$ (Fig. \ref{figprofile}). As such, during the first 20 time-steps, $A$'s relatively poor-in-comparison self-reputation has an immediate negative effect on the \textit{value} it associates with the \texttt{trade\_with\_B} action during the look-ahead search. It quickly becomes more \textit{valuable} to stop trading with $B$. 
Similarly, $A$'s reputation needs to be high for it to start trading with $B$ again, explaining the slow increase of the frequency of trade offers at the start of the second phase.

\subsection{Experiment 2: Trading between three agents} \label{ex:trade3ag}
Let $A$, $B$, and $C$ be three agents. Each agent simultaneously plays the role of buyer and seller, and can thus engage in a trade with any other agent. Each agent can accept or refuse any trade offer made by any remaining agent.

The present scenario is used to verify the ability of a RepNet agent, say agent $A$, to manage its trades with the two remaining agents $B$ and $C$, based not only on their behavior towards the agent of interest but also their behavior with each other.

In the first part, agent $B$ is asked to refuse each trade offer made by agent $A$, while agent $C$ is expected to accept each trade offer coming from $A$. This portion of the experiments assesses the ability of the RepNet agent (agent $A$) to accurately determine which agent it is more likely to successfully engage in trades with. In the second part, the roles are switched, and agent $B$ accepts the trade offers, while agent $C$ refuses them. This portion assesses the ability of the agent of interest to \textit{unlearn} what it has learned and adapt its behavior accordingly. 
In the third and final part, the RepNet agent is asked to not trade with either $B$ or $C$, that is, to only make use of the \texttt{wait} action. Said differently, the optimal action according to its planning, while tracked throughout the experiment, is not performed on the environment. All the while, agents $B$ and $C$ are asked to engage in trades with each other. Agent $B$ is asked to reject all trade offers, while agent $C$ is asked to accept all trade offers. 
The variables tracked are the action distribution and reputation of $B$ and $C$ \textit{in the eyes of agent $A$}, as well as the evolution of whom agent $A$ would rather trade with. 
This portion of the experiment aims at testing the ability of the RepNet agent to draw conclusions on how it should act based on interactions it is not directly affected by.

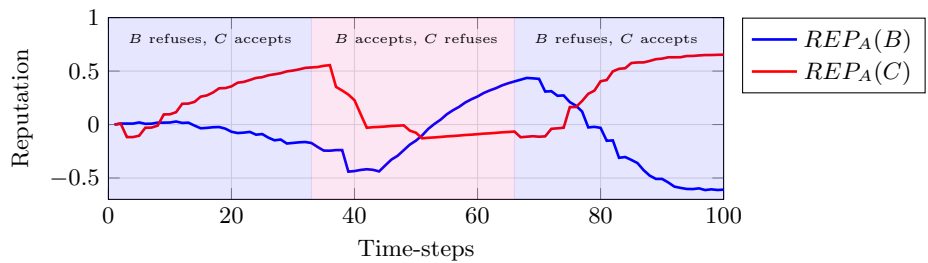
\begin{figure}
    \centering
\begin{tikzpicture}
\begin{axis}[
	xlabel=Time-steps,
	ylabel= Reputation,
	title style={align=left},
	grid=both,
	minor grid style={gray!25},
	major grid style={gray!25},
	legend pos=outer north east,
	width=0.8\linewidth,
	height=4cm,
	no marks,
	xmin=0,
    xmax=100,
    ymin=-0.7,
    ymax=1]
\addplot[line width=1pt,solid,color=blue] %
	table[x=b,y=a,col sep=comma]{tradeABCrepB.csv};
\addlegendentry{$REP_A(B)$};

\addplot[line width=1pt,solid,color=red] %
	table[x=b,y=a,col sep=comma]{tradeABCrepC.csv};
\addlegendentry{$REP_A(C)$};


\addplot+[draw=blue,fill=blue, opacity=0.1]
        coordinates {(0,-1) (0,1.1) (33,1.1) (33,-1)};

\addplot+[draw=magenta,fill=magenta, opacity=0.1]
        coordinates {(33,-1) (33,1.1) (66,1.1) (66,-1)};
        
\addplot+[draw=blue,fill=blue, opacity=0.1]
        coordinates {(66,-1) (66,1.1) (100,1.1) (100,-1)};

\node[] at (16.5,32pt) {\tiny{$B$ refuses, $C$ accepts}};    
\node[] at (50,32pt) {\tiny{$B$ accepts, $C$ refuses}};    
\node[] at (82.5,32pt) {\tiny{$B$ refuses, $C$ accepts}};  

\end{axis}
\end{tikzpicture}
    \caption{Reputation of agents $B$ and $C$, according to $A$}
    \label{fig4}
\end{figure}

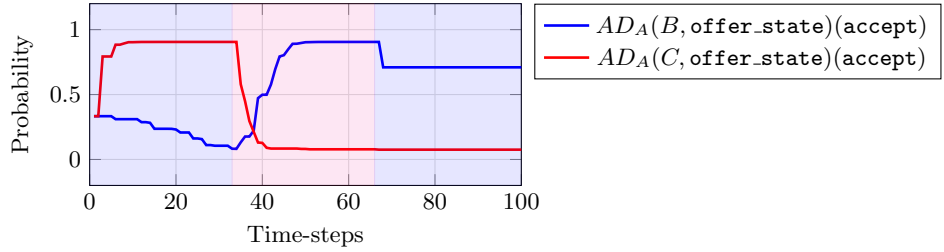
\begin{figure}
    \centering
\begin{tikzpicture}
\begin{axis}[
	xlabel=Time-steps,
	ylabel= Probability,
	title style={align=left},
	grid=both,
	minor grid style={gray!25},
	major grid style={gray!25},
	legend pos=outer north east,
	width=0.6\linewidth,
	height=4cm,
	no marks,
	xmin=0,
    xmax=100,
    ymin=-0.2,
    ymax=1.2]
\addplot[line width=1pt,solid,color=blue] %
	table[x=b,y=a,col sep=comma]{tradeABCacceptB.csv};
\addlegendentry{$AD_A(B,\texttt{offer\_state})(\texttt{accept})$};

\addplot[line width=1pt,solid,color=red] %
	table[x=b,y=a,col sep=comma]{tradeABCacceptC.csv};
\addlegendentry{$AD_A(C,\texttt{offer\_state})(\texttt{accept})$};


\addplot+[draw=blue,fill=blue, opacity=0.1]
        coordinates {(0,-1) (0,1.5) (33,1.5) (33,-1)};

\addplot+[draw=magenta,fill=magenta, opacity=0.1]
        coordinates {(33,-1) (33,1.5) (66,1.5) (66,-1)};
        
\addplot+[draw=blue,fill=blue, opacity=0.1]
        coordinates {(66,-1) (66,1.5) (100,1.5) (100,-1)};
\end{axis}
\end{tikzpicture}
    \caption{Probability of agents $B$ and $C$ accepting trade offers from agent $A$, according to $A$}
    \label{fig5}
\end{figure}

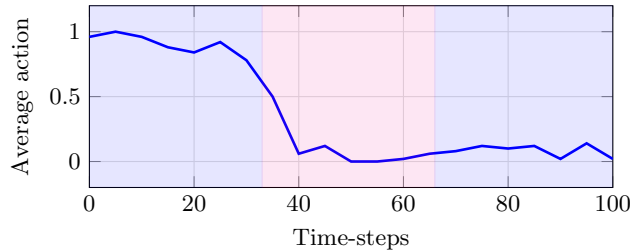
\begin{figure}
\centering
\begin{tikzpicture}
\begin{axis}[
	xlabel=Time-steps,
	ylabel= {Average action},
	grid=both,
	minor grid style={gray!25},
	major grid style={gray!25},
	legend pos=outer north east,
	width=0.7\linewidth,
	height=4cm,
	no marks,
	xmin=0,
    xmax=100,
    ymin=-0.2,
    ymax=1.2]
\addplot+[line width=1pt,solid,color=blue] plot coordinates { (0, 0.96) (5, 1) (10, 0.96) (15, 0.88) (20, 0.84) (25,0.92)(30, 0.78)(35, 0.5)(40, 0.06)(45, 0.12)(50, 0)(55, 0)(60, 0.02)(65, 0.06)(70, 0.08)(75, 0.12)(80, 0.1)(85, 0.12)(90, 0.02)(95, 0.14)(100, 0.02)};



\addplot+[draw=blue,fill=blue, opacity=0.1]
        coordinates {(0,-1) (0,1.5) (33,1.5) (33,-1)};

\addplot+[draw=magenta,fill=magenta, opacity=0.1]
        coordinates {(33,-1) (33,1.5) (66,1.5) (66,-1)};
        
\addplot+[draw=blue,fill=blue, opacity=0.1]
        coordinates {(66,-1) (66,1.5) (100,1.5) (100,-1)};
\end{axis}
\end{tikzpicture}
\caption{Average action taken by agent $A$. Action $a = 0$ corresponds to trading with agent $B$, action $a = 1$ corresponds to trading with agent $C$.}
\label{fig6}
\end{figure}

Fig. \ref{fig4} shows the evolution of the reputations of agents $B$ and $C$. Fig. \ref{fig5} displays the evolution of the probabilities of agents $B$ and $C$ accepting trade offers from agent $A$. Finally, Fig. \ref{fig6} shows the
evolution of whom agent $A$ would rather trade with.

Agent $B$ is told to refuse, and agent $C$ to accept, each trade offer during the first 33 time-steps. In accordance with the results obtained in Section \ref{ex:trade2ag}, agent $A$ is able to pick up on the other agents' behavioral habits it is affected by. As a result, the reputation of $B$ and its probability of accepting trade offers decrease. Similarly, the reputation of $C$ and its probability of accepting trade offers increase. All the while, agent $A$ chooses to conduct the majority of its trades with $C$.
The following 33 time-steps reverse $B$'s and $C$'s roles. Similarly, agent $A$ is able to adapt its behavior accordingly and ends up trading mostly with $B$. The reputation of $B$ has increased, while the reputation of $C$ has decreased.

During the last 33 time-steps, agents $B$ and $C$ are tasked with trading with one another while $A$ plays the role of observer, that is, only makes use of the \texttt{wait} action. $B$ is asked to refuse all trade offers, while $C$ is asked to accept all trade offers. 
Interestingly, Fig. \ref{fig6} shows that, based on its planning, agent $A$ would prefer to keep trading with $B$, even though the reputation of $B$ decreases and the reputation of $C$ increases in the eyes of $A$. Said differently, as long as $B$ does not refuse $A$'s offers directly, agent $A$ will prefer to trade with $B$ over $C$.

The explanation for this is twofold. Firstly, the subjective transition probability of a trade $A$ might want to do with $B$ is, in the eyes of $A$, conditioned only by $A$'s \textit{own} reputation. As such, $B$'s falling or rising reputation has no bearing on $A$'s decision-making. Secondly, the probability of $B$ accepting (or refusing) $A$'s trade offer, according to $A$, can only be updated through the direct experience it has with $B$. As such, the action distribution does not change and can thus not influence $A$ decision-making either. 

The simplest way of alleviating this shortcoming is to extend the subjective transition model. Adding the reputation of the agent at the receiving end of the trade offer (e.g., agent $B$) as a parameter to the subjective transition model would allow agent $A$ to incorporate other agents' reputation in its decision-making process. As such, if the subjective transition probability of $B$ accepting $A$'s trade offer were given by
\begin{align*}
   ST_A(A, \texttt{offer\_state}, \texttt{wait\_s}, \texttt{accept\_state}, r_A, \textcolor{cyan}{r_B}) ,
\end{align*}
where the newly introduced parameter $\textcolor{cyan}{r_B}$ is $B$'s reputation, agent $A$ could make use of $\textcolor{cyan}{r_B}$ to assist with its decision-making. This comes with the drawback of increasing the complexity of designing the subjective transition model.

\section{Summary and future work}

In this paper, we revised the multi-agent framework called RepNet introduced by \textit{Rens et al.} \cite{rensetal}, addressed its mathematical inconsistencies and proposed a online learning algorithm for finding approximate solutions. The viability of the framework was then tested in a series of experiments.

The current definition of \textit{objective} transitions could be extended to incorporate the reputation of agents other than the RepNet agent. The experimental results showed that the RepNet agent is incapable of adapting its behavior to situations that do not directly affect it. 
Including the reputation of the agent at the receiving end of a directed action in the \textit{directed} transition model
is likely to lead to better-informed decision-making.

We did not address partially observable environments. Many real-world problems do not benefit from full observability, bringing the updated RepNet framework back to a partially observable setting should be considered for future work.

The small-scale experiments conducted in Section \ref{sec:exp} served as a proof of concept for the RepNet framework. While applying the framework to problems of realistic size was beyond the scope of this paper, the absence of large-scale tests does raise questions as to the scalability of this approach.
Real-world problems can easily become too complex for transition models to be designed by any one person without leveraging common state features \cite{bout2}. 
A compact way to represent real-world state spaces can be achieved by introducing elements of relational logic \cite{bout2}.
From a \textit{logic programming} point of view, a state space is hereby defined by a collection of relations, while a state is an \textit{interpretation} of this collection \cite{relationallogic}. Transition models and reward schemes are then represented by \textit{probabilistic rules} \cite{Nitti2017}.

\bibliographystyle{splncs04}
\bibliography{paper}

\end{document}